\documentclass[lettersize,journal]{IEEEtran}
\usepackage{amssymb,amsmath}
\usepackage{diagbox}
\usepackage{multirow}
\usepackage[normalem]{ulem}
\usepackage{booktabs}
\usepackage{algorithmic}
\usepackage{algorithm}
\usepackage{array}
\usepackage[caption=false,font=normalsize,labelfont=sf,textfont=sf]{subfig}
\usepackage{textcomp}
\usepackage{stfloats}
\usepackage{url}
\usepackage{verbatim}
\usepackage{graphicx}
\usepackage{cite}
\usepackage{wrapfig}
\usepackage{url}
\usepackage{color}
\usepackage[justification=centering]{caption}
\begin{document}

\title{Cellular Traffic Prediction via Deep State Space Models with Attention Mechanism}
\author{Hui~Ma,       
         Kai~Yang,\IEEEmembership{Senior Member, IEEE}, and Man-On~Pun, \IEEEmembership{Member, IEEE}
\thanks{This work was supported in part by National Natural Science Foundation of China under Grant 61771013. (Corresponding author: Kai Yang)}
\thanks{Hui Ma and Kai Yang are with the College of Electronic and Information Engineering, Tongji University, Shanghai 201804, China. (e-mail: kaiyang@tongji.edu.cn). }
\thanks{Man-On Pun is with the School of Science and Engineering, The Chinese University of Hong Kong at Shenzhen, Shenzhen 518172, China, also with the Shenzhen Key Laboratory of IoT Intelligent Systems and Wireless Network Technology, Chinese University of Hong Kong at Shenzhen, Shenzhen 518172, China, and also with the Shenzhen Research Institute of Big Data, Chinese University of Hong Kong at Shenzhen, Shenzhen 518172, China (e-mail: simonpun@cuhk.edu.cn).}}

\maketitle
\begin{abstract} 
Cellular traffic prediction is of great importance for operators to manage network resources and make decisions. Traffic is highly dynamic and influenced by many exogenous factors, which would lead to the degradation of traffic prediction accuracy. This paper proposes an end-to-end framework with two variants to explicitly characterize the spatiotemporal patterns of cellular traffic among neighboring cells. It uses convolutional neural networks with an attention mechanism to capture the spatial dynamics and Kalman filter for temporal modelling. Besides, we can fully exploit the auxiliary information such as social activities to improve prediction performance. We conduct extensive experiments on three real-world datasets. The results show that our proposed models outperform the state-of-the-art machine learning techniques in terms of prediction accuracy.
\end{abstract}


\section{Introduction}
\IEEEPARstart{W}{ith} the development of cellular mobile communications and the popularization of various mobile devices and applications, there is a proliferation of cellular users. According to a Cisco network report\cite{Cisco}, there will be 5.7 billion cellular network users worldwide by 2023, accounting for 71\% of the global population. In addition, with the deployment and implementation of the fifth generation (5G) wireless systems, a growing number of Internet of things (IoT) devices, a high Internet bandwidth, and fast network speeds will contribute to traffic explosion\cite{b02,b2}. Besides, the growing scale and complexity of the Internet pose significant challenges for network management. Researchers introduce artificial intelligence technologies and build prediction models with machine learning methods to accurately estimate network traffic. For instance, the international telecommunication union (ITU) has established the Focus Group as a platform for bringing more automation and intelligence to network design and management\cite{b3,DNA,AIIOT}. Moreover, accurate traffic prediction plays an essential role in enhancing the quality of service (QoS)\cite{b4}, alleviating network congestions, and reducing operational expenditures\cite{b5}.

Existing cellular traffic prediction methods face the following challenges. \textit{Firstly}, cellular traffic is highly dynamic and influenced by many exogenous factors. For example, important news events or large-scale sports activities may cause a surge in traffic value to a certain extent. Exploiting auxiliary information to infer highly time-varying traffic patterns and improve the prediction performance is a challenge in our study. \textit{Secondly}, cellular traffic data have complex spatiotemporal correlations among adjacent cells\cite{b6, b7}. On the one hand, Internet traffic follows periodic patterns over weekdays (e.g., traffic demand is high during the daytime and
low in the nighttime). On the other hand, user mobilities make network traffic fluctuate significantly and contribute to spatial correlations of cellular traffic generated by neighboring base stations (BSs). In this case, a challenge worth studying is how to effectively extract complex spatiotemporal dynamics and build a spatiotemporal prediction model with high accuracy.

In recent years, there has been a rapid growth in applying deep learning approaches to characterize complex spatiotemporal correlations. In particular, recurrent neural network (RNN) based models such as long short-term memory (LSTM) neural networks and gate recurrent unit (GRU) neural networks have been widely utilized to capture the long-term temporal dependencies\cite{b16,b17,b18,PC2A}. However, these methods are inefficient and time consuming for the long-term forecasting\cite{add-ref3}. To tackle the shortcomings of RNN based models, we incorporate Kalman filter with low complexity and interpretability in our study. In Kalman filter\cite{Zhao2019_iccv}, the optimal posteriori state can be corrected by incorporating a new observation into the priori state at each timestep. Thus, approximation errors can be minimized over time, providing more accurate predictions. Nonetheless, Kalman filter is still insufficient because it is challenging to model the spatial correlations of traffic sequences. This prompts us to consider deep neural networks, where convolutional neural networks (CNN) are distinguished by their superior performance for modeling grid-based spatial dependencies of cellular traffic data. In this way, the complex spatiotemporal dependencies can be captured in our study.

Our study proposes a novel deep state space model (DSSM) to perform cellular traffic forecasting. The contributions are three-fold. \textit{Firstly}, since auxiliary information helps to infer highly time-varying dynamics, we design a traffic prediction model based on multi-source heterogeneous data (including structural data from telco datasets and textual data from social pulse and website news) to improve prediction accuracy. \textit{Secondly}, we propose an end-to-end framework for spatiotemporal traffic prediction that implements Kalman filter and attention-based CNN to model the long-term temporal and spatial dynamics, respectively. In particular, the Kalman gain in DSSM is an adaptive weight that can proportionally correct the posteriori state by combining the observation and priori state in every timestep, thereby ameliorating error accumulation over time. \textit{Thirdly}, the changes in the Kalman gain can reflect the importance of the observation and priori state at each timestep to predictions. The visualization of the Kalman gain will help us better understand how the system state changes over time when capturing the long-term temporal dependencies, thereby making the model partially interpretable.
  
Furthermore, the linear Kalman filter (LKF) is an optimal linear quadratic estimator in linear systems. In contrast, the extended Kalman filter (EKF) can approximate nonlinear functions through Taylor series expansion and achieve optimal state estimation in nonlinear systems. Based on LKF and EKF, we designed two variants including an attention-based linear Kalman filter (A-LKF) and an attention-based extended Kalman filter (A-EKF). 

Extensive experiments have been carried out on three real-world datasets to evaluate the prediction accuracy of our proposed A-LKF and A-EKF. It is revealed that the proposed methods exhibit strong empirical performance and outperform other state-of-the-art techniques.

The remainder of this paper is organized as follows. Section II provides a brief introduction of related work. Section III describes the problem formulation and the details of our proposed models, including A-LKF and A-EKF. Section IV presents the detailed experiment setup, followed by the analysis of the results in Section V. Finally, we conclude our study in Section VI.

\section{Related Work}
The cellular traffic prediction methods can be roughly classified into two categories, including statistical and machine learning approaches\cite{b02}. The classical statistical models have been widely investigated, such as autoregression (AR), autoregressive integrated moving average\cite{b9}, exponential smoothing\cite{b10}, and state space model (SSM)\cite{Chen2021a_DynaNet,Revach2022_KalmanNet,b12}. These traditional methods are usually interpretable but often perform poorly in traffic prediction. 
 
Machine learning techniques can be divided into traditional and deep learning approaches. The traditional machine learning methods, such as vector autoregression and support vector regression\cite{b14}, maintain excellent prediction performance compared with statistical approaches and are extensively used for cellular traffic forecasting. Besides, various studies have also assessed the availability of deep learning techniques for cellular traffic prediction\cite{Abbasi2021,mao2018deep,JIANG2022117163}. 
The main advantages of deep learning approaches are handling a large amount of data and modeling traffic dynamics effectively.

RNN and variants such as LSTM neural networks and GRU neural networks can effectively capture the temporal dependencies of network traffic and have a wide range of applications for cellular traffic prediction\cite{qin2017a,b15,Kongitsg-2019}. For instance, Shiang \textit{et al}.\cite{b17} compared the prediction performance between GRU and LSTM neural networks. The results showed that the convergence speed of GRU neural networks is faster than that of LSTM neural networks because GRU neural networks are less complex with a smaller number of gates. Trinh \textit{et al}. \cite{b18} studied stacked multiple LSTM neural networks to capture temporal dependencies of mobile traffic, which verified the effectiveness of LSTM neural networks for multistep prediction. Gao \textit{et al}. \cite{b19} proposed a spatiotemporal attention architecture with hierarchical LSTM neural networks to extract meaningful information. Wang \textit{et al}. \cite{b20} extracted the periodic patterns of cellular traffic through Fourier analysis, and modeled the long-term dependencies and remaining random components with LSTM neural networks and Gaussian process regression, respectively. Wang \textit{et al}. \cite{Wang2021a-3009159} proposed a data-augmentation-based cellular traffic prediction model. The generative adversarial networks were utilized to generate training samples, which aims to solve the insufficiency of training data. Moreover, LSTM neural networks were used to achieve a multistep prediction. However, the above methods have difficulty in characterizing traffic's spatial dependencies among adjacent cells. 

There is a growing body of literature that graph convolutional networks (GCN)\cite{b21,9158437,He2022_gcn,Jiang2021-graph-based-survey}, autoencoder, and CNN are employed to explicitly characterize the spatial dependencies for traffic demand forecasting\cite{8845132,7542585}. For example, Wang \textit{et al}.\cite{b22} exploited graph neural networks to model the spatiotemporal correlations of urban cellular traffic. However, the graph-based deep learning approaches suffer several drawbacks, such as an enormous computation burden. In terms of autoencoder-based models, Chen \textit{et al}.\cite{Chen2020-2953745} proposed a sparse autoencoder to capture useful features from high-dimensional workload data and used GRU neural networks to extract the long-term temporal dependencies. In \cite{b23}, Jing \textit{et al}. utilized a stacked autoencoder to capture the spatial relationships in a Mobile dataset. However, it is not easy to achieve high prediction accuracy. In \cite{b25}, Zhang \textit{et al}. modeled the spatiotemporal dependencies by adopting convolutional blocks, indicating the effectiveness of CNN for mobile traffic prediction. Liu \textit{et al}. \cite{b26} categorized traffic volumes into three groups: hourly, daily, and weekly data. Then, three paralleled CNN with residual connections were employed to extract cellular traffic's closeness, periodicity, and trend patterns. Huang \textit{et al}. \cite{b27} investigated three deep learning models, including CNN, RNN, and a combination of CNN and RNN (CNN-RNN) for traffic forecasting. The experiments revealed that CNN-RNN performed best in modeling spatial and temporal dependencies. Du \textit{et al}. \cite{Du2020-2900481} proposed an irregular CNN to extract the spatial dependencies and used LSTM neural networks to model the temporal relationships. Besides, three paralleled structures were employed to characterize the recent, daily, and weekly traffic passenger flows, respectively. Shen \textit{et al}. \cite{add-ref3} designed a time-wise attention mechanism to capture the long-range temporal relationships and used CNN to capture the spatial dependencies. Zhao \textit{et al}. \cite{Zhaonan-2020} proposed a spatial-temporal attention-convolution network with GCN and CNN to capture the spatial and temporal features simultaneously. Liu \textit{et al}. \cite{add-ref2} designed ST-Tran to explore the temporal and spatial sequences with transformer. Moreover, Wang \textit{et al}. \cite{Wang2020a-3025580} provided a comprehensive review of deep learning models for spatiotemporal data mining.
 
Furthermore, various exogenous factors are considered in some studies for cellular traffic prediction\cite{DeepTP}. For instance, Assem \textit{et al}. \cite{b29} proposed ST-DenNetFus with a late fusion method and introduced the day of the week, functional regions, and crowd mobilities. Zhang \textit{et al}. \cite{b31} considered BSs, point of interest (PoI), and social activities of a cell for cellular traffic forecasting. Bhorkar \textit{et al}. \cite{b32} investigated network configurations and other auxiliary factors for cell load prediction. Besides, Du \textit{et al}. \cite{Du2020-2900481} explored the impact of external information such as weather conditions and social events on the urban traffic passenger flows. Zhao \textit{et al}.\cite{Zhaonan-2020} considered traffic data from neighboring cells, PoI, and weather situations, while Zhao \textit{et al}.\cite{9158437} considered user movement patterns as external factors. Moreover, Yao \textit{et al}. \cite{Yaoweiran-sq-2021} investigated that Twitter messages help better understand the impact of people's activity patterns on urban traffic and thus improve the prediction performance. Collectively, these studies outline the critical role of auxiliary information in cellular traffic prediction. 

\section{Problem Formulation and Framework}
In this section, we first briefly present the problem formulation of cellular traffic prediction and then discuss the details of the proposed models (i.e., A-LKF and A-EKF) in the following sections.

\subsection{Problem Formulation}
In this study, we consider the high-dimensional time series forecasting problem. Let $\boldsymbol{X}\in \mathbb{R}^{T \times D}=[\boldsymbol{x}_1,\boldsymbol{x}_2,...,\boldsymbol{x}_T]^{\top}$ denote cellular traffic data. Likewise, let $\boldsymbol{E}\in \mathbb{R}^{T \times {D_e}}=[\boldsymbol{e}_1,\boldsymbol{e}_2,...,\boldsymbol{e}_T]^{\top}$ denote external data. 
Our study mainly considers two sources of auxiliary information, including social activities and news. Regarding social activities, we take the location information, the number of users and tweets into consideration. Regarding news data, we consider the number and type of articles published over a period as well as other latent attributes such as holidays and the day of the week.

The formulation of our study can be expressed as the $h$-step-ahead forecast problem. To be specific, we aim at learning a predictor $g(\cdot)$ to estimate the traffic volume $\hat{\boldsymbol{Y}}\in \mathbb{R}^{h \times D}=[\boldsymbol{y}_{T+1},\boldsymbol{y}_{T+2},...,\boldsymbol{y}_{T+h}]^{\top}$, which is shown as follows,
\begin{equation}
\hat{\boldsymbol{Y}}=g(\boldsymbol{x}_{1},\boldsymbol{x}_{2},...,\boldsymbol{x}_{T},\boldsymbol{e}_{1},\boldsymbol{e}_{2},...,\boldsymbol{e}_T), 
\end{equation}
where ${h}$ denotes the prediction horizon ahead of the current timestep. In a cellular traffic scenario, ${h}$ usually ranges from hours to days.

\subsection{Framework}
Fig. 1 demonstrates an overview of our proposed DSSM with five significant modules, including the exogenous feature extraction, an encoder, Kalman filter, a decoder, and AR. Firstly, the exogenous feature extraction aims to achieve a uniform representation of compelling auxiliary features, and AR is responsible for capturing the scale variations of cellular traffic. Besides, the encoder module is implemented with CNN and an attention mechanism to capture the spatial dependencies of cellular traffic among adjacent cells, whereas Kalman filter is utilized to estimate the optimal state. Furthermore, the outputs of AR, Kalman filter, and exogenous feature extraction are fused to obtain the final output. We will introduce the building blocks in detail. 

\begin{figure}[htb]
\begin{center}
    \centerline{\includegraphics[width=\columnwidth]{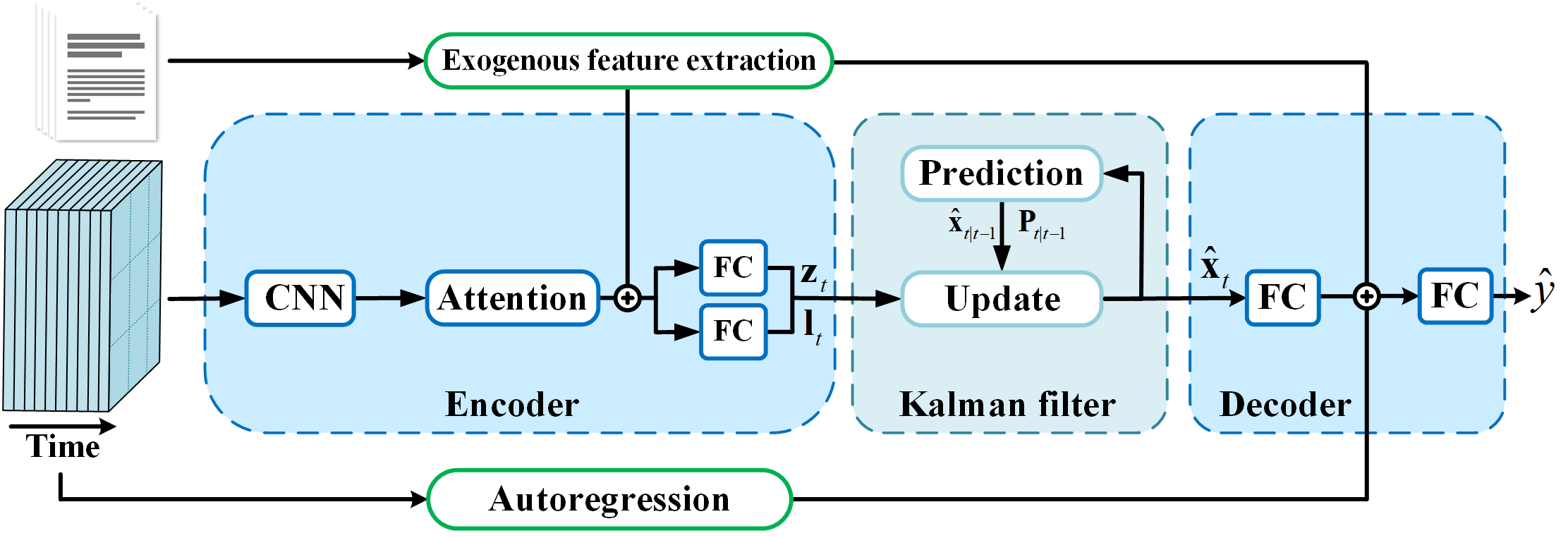}}
Fig. 1. The structure of the deep state space model.
\label{Fig}
\end{center}
\end{figure}

\textbf {Exogenous feature extraction}:
The exogenous feature extraction is designed to encode auxiliary information including social activities, news, and other latent attributes into a feature vector. This paper uses the fully connected layer (FC) to obtain $\boldsymbol{O}^{e_1}_t\in \mathbb{R}^{T\times D_a}=[\boldsymbol{o}^{e_1}_1,\boldsymbol{o}^{e_1}_2,...,\boldsymbol{o}^{e_1}_T]^{\top}$ and $\boldsymbol{O}^{e_{2}}_t\in \mathbb{R}^{T\times D}=[\boldsymbol{o}^{e_2}_1,\boldsymbol{o}^{e_2}_2,...,\boldsymbol{o}^{e_2}_T]^{\top}$, which will be further used in the encoder and decoder module, respectively.

\textbf {Autoregression}:
As mentioned in \cite{b37}, deep learning models such as RNN can effectively capture the existing patterns from training data but have difficulty extracting insight from test data. To alleviate this problem, a linear model is employed to capture the scale changes from the test data and improve the prediction performance, which has also been discussed in \cite{b38}. In this paper, we utilize AR as the linear module. Let $\boldsymbol{W}_{ar}$ and $\boldsymbol{b}_{ar}$ denote weight and bias respectively, then the output $\boldsymbol{o}^{ar}_{t}\in \mathbb{R}^{D}$ at time $t$ can be calculated by using (2),
\begin{equation}
\boldsymbol{o}^{ar}_{t}= \boldsymbol{W}_{ar}^{\top} \boldsymbol{x}_t + \boldsymbol{b}_{ar}.
\end{equation}

\textbf {Encoder}: The encoder module utilizes CNN and an attention mechanism to extract the spatial features of cellular traffic. With the ability to model local dependencies among variables, the convolutional blocks are adopted to capture the spatial characteristics via the following formulation,
\begin{equation}
\boldsymbol{O}^{cnn}=g_{cnn}(\boldsymbol{x}_{1},\boldsymbol{x}_{2},...,\boldsymbol{x}_{T}), 
\end{equation}
where $\boldsymbol{O}^{cnn}\in \mathbb{R}^{T \times D_c}=[\boldsymbol{o}^{cnn}_{1}, \boldsymbol{o}^{cnn}_{2},\cdots,\boldsymbol{o}^{cnn}_{T}]^{\top}$. $g_{cnn}(\cdot)$ represents the convolutional operation with layer normalization\cite{b33} followed by Relu function as the activation function. 

Besides, we employ a self-attention mechanism to select relevant hidden states across all timesteps automatically. Specifically, the query, key, and value matrices are denoted as $\boldsymbol{Q},\boldsymbol{K},$ and $\boldsymbol{V}$, respectively. They can be obtained via matrix multiplication between $\boldsymbol{O}^{cnn}$ and learnable weights, namely $\boldsymbol{W}^{Q},\boldsymbol{W}^{K},$ and $\mathbf{W}^{V}$,
\begin{equation}
\boldsymbol{Q}=\boldsymbol{O}^{cnn} \boldsymbol{W}^{Q},  
\boldsymbol{K}=\boldsymbol{O}^{cnn} \boldsymbol{W}^{K},
\boldsymbol{V}=\boldsymbol{O}^{cnn} \boldsymbol{W}^{V}.
\end{equation}

The output of the self-attention mechanism $\boldsymbol{o}^{att}_{t} \in \mathbb{R}^{{D_a}}$ can be computed via a scaled dot-product function\cite{b33} by using (5),
\begin{equation}
\boldsymbol{o}^{att}_{t} = softmax(\frac{\boldsymbol{QK}^{\top}}{\sqrt{d_c}})\boldsymbol{V}.
\end{equation}

Then, $\boldsymbol{o}^{att}_{t}$ and $\boldsymbol{o}^{e_1}_t$ are aggregated as $\boldsymbol{o}^{concat}_{t}\in \mathbb{R}^{D_a}$. The two paralleled FC output an observation $\boldsymbol{z}_t \in \mathbb{R}^{D_k}$ and a vector $\boldsymbol{l}_t \in \mathbb{R}^{D_k}$ with the following formula,
\begin{eqnarray}
\boldsymbol{z}_t &=& Relu(\boldsymbol{W}_{k_{1}} \boldsymbol{o}^{concat}_{t}+\boldsymbol{b}_{k_{1}}), \\
\boldsymbol{l}_t &=& Relu(\boldsymbol{W}_{k_{2}}  \boldsymbol{o}^{concat}_{t}+\boldsymbol{b}_{k_{2}}), 
\end{eqnarray}
where $\boldsymbol{z}_t$ is a new observation, which can be used to estimate the posteriori state while $\boldsymbol{l}_t$ is used to calculate the measurement noise $\boldsymbol{R}_{t}$ in Kalman filter module. $\boldsymbol{W}_{k_{1}},\boldsymbol{W}_{k_{2}},\boldsymbol{b}_{k_{1}},\boldsymbol{b}_{k_{2}}$ represent the learnable parameters, respectively. 

\textbf {Kalman Filter}: 
As a famous algorithm in SSM, Kalman filter\cite{b34} can achieve the optimal estimation with state equations and measurement equations, which has been widely used in dynamic systems. 

In Kalman filter\cite{b34,b35,2019Moving}, it is assumed that the process noise and the measurement noise are zero-mean Gaussian random variables with covariance matrices $\boldsymbol{Q}$ and $\boldsymbol{R}_{t}$, respectively. $\boldsymbol{Q}$, $\boldsymbol{R}_{t}$, and estimated state covariance matrix $\boldsymbol{P}_{t}$ are assumed as diagonal matrices in our study. Let $\boldsymbol{l}_t$ and $\boldsymbol{\lambda}$  denote the eigenvector of $\boldsymbol{R}_t$ and $\boldsymbol{Q}$, respectively. $\boldsymbol{l}_t$ can be calculated via equation (7) and $\boldsymbol{\lambda}$ can also be obtained via the FC. Then, $\boldsymbol{R}_t$ can be achieved via $\boldsymbol{R_{t}}$=diag($\boldsymbol{l}_{t}$), while $\boldsymbol{Q}$ can be obtained by $\boldsymbol{Q}$=diag($\boldsymbol{\lambda}$). 

In our study, we propose A-LKF and A-EKF based on linear Kalman filter and extended Kalman filter, respectively. In A-LKF, the estimated optimal state can be calculated with the following two steps\cite{b34}. 

\begin{itemize}
\item Prediction step: Specifically, let $\hat{\boldsymbol{x}}_{t-1}$ and $\boldsymbol{P}_{t-1}$ denote the optimal posteriori state and covariance at ${t-1}$ timestep, respectively. Then the priori state $\hat{\boldsymbol{x}}_{t \mid t-1}$ and covariance $\boldsymbol{P}_{t \mid t-1}$ at time $t$ can be computed by using (8) and (9),
\begin{eqnarray}
\hat{\boldsymbol{x}}_{t \mid t-1}&=&\boldsymbol{F}\hat{\boldsymbol{x}}_{t-1},\\
\boldsymbol{P}_{t \mid t-1}&=&\boldsymbol{F}\boldsymbol{P}_{t-1} \boldsymbol{F}^{\top}+ \boldsymbol{Q},
\end{eqnarray}
where $\boldsymbol{F}$ represents the transition matrix. It is assumed as a diagonal matrix with the eigenvector of $\boldsymbol{\gamma}$, which can be learned from a neural network. Then, $\boldsymbol{F}$ can be obtained via $\boldsymbol{F}$=diag($\boldsymbol{\gamma}$).

\item Update step: The new observation $\boldsymbol{z}_t$ is available at time $t$ by using (6) and the Kalman gain, represented as $\boldsymbol{K}$, is computed by using (10). Then the optimal posteriori state $\hat{\boldsymbol{x}}_{t}$ and covariance $\boldsymbol{P}_{t}$ at time ${t}$ can be be obtained via the measurement update equations\cite{2011Kalman}, which are given as follows, 
\begin{align}
\boldsymbol{K}&=\boldsymbol{P}_{t \mid t-1}\boldsymbol{H}^{\top} (\boldsymbol{HP}_{t \mid t-1}\boldsymbol{H}^{\top} + \boldsymbol{R}_{t})^{-1},\\
\hat{\boldsymbol{x}}_{t}&=\hat{\boldsymbol{x}}_{t \mid t-1}+\boldsymbol{K}(\boldsymbol{z}_{t}-\boldsymbol{H}\hat{\boldsymbol{x}}_{t \mid t-1}),\\
&=\boldsymbol{K}\boldsymbol{z}_{t}+(\boldsymbol{I} - \boldsymbol{K}\boldsymbol{H})\hat{\boldsymbol{x}}_{t \mid t-1},\\
\boldsymbol{P}_{t}&=(\boldsymbol{I} - \boldsymbol{K}\boldsymbol{H})\boldsymbol{P}_{t \mid t-1},
\end{align}
where $\boldsymbol{I}$ denote the identity matrix. Please note the matrix ${\boldsymbol{H}}$ is an identity matrix in A-LKF. 

\quad It can be observed from Equation (12) that the Kalman gain is an adaptive weight to balance the observation $\boldsymbol{z}_t$ and the priori state $\hat{\boldsymbol{x}}_{t \mid t-1}$. A larger Kalman gain represents a more significant contribution of the observation to correct the posteriori state and vice versa.
\end{itemize}

Likewise, A-EKF can also be conducted with the prediction step and update step to achieve the optimal state estimation\cite{b36}.

\begin{itemize}
\item Prediction step: The priori state $\hat{\boldsymbol{x}}_{t \mid t-1}$ and covariance $\boldsymbol{P}_{t \mid t-1}$ can be obtained by using (14) and (15),
\begin{eqnarray}
\hat{\boldsymbol{x}}_{t \mid t-1}&=&f(\hat{\boldsymbol{x}}_{t-1}),\\
\boldsymbol{P}_{t \mid t-1}&=&\boldsymbol{J}_{f}(\hat{\boldsymbol{x}}_{t-1})\boldsymbol{P}_{t-1} \boldsymbol{J}_{f}^{\top}(\hat{\boldsymbol{x}}_{t-1})+\boldsymbol{Q},
\end{eqnarray}
where $f(\cdot)$ denotes a nonlinear function and $\boldsymbol{J}_{f}(\cdot)$ is the Jacobian of $f(\cdot)$. Many types of nonlinear functions can be used in our study. The quadratic function is a nonlinear function that has been widely used in EKF\cite{roth2013kalman,skoglund2019iterative}. Therefore, we use $f(x)=\alpha_0 + \alpha_1 x + \alpha_2 x^2$ as the nonlinear function in A-EKF, where $\alpha_0,\alpha_1,\alpha_2$ can be obtained via a neural network.

\item Update step: The optimal posteriori state $\hat{\boldsymbol{x}}_{t}$ and covariance $\boldsymbol{P}_{t}$ at time $t$ can be calculated with the following formula, 
\begin{eqnarray}
\boldsymbol{S}_{t} &=& \boldsymbol{J}_{h}(\hat{\boldsymbol{x}}_{t \mid t-1}) \boldsymbol{P}_{t \mid t-1}\boldsymbol{J}_{h}^{\top}(\hat{\boldsymbol{x}}_{t \mid t-1})+\boldsymbol{R}_{t}, \\
\boldsymbol{K} &=& \boldsymbol{P}_{t \mid t-1}\boldsymbol{J}^{\top}_{h}(\hat{\boldsymbol{x}}_{t \mid t-1}) \boldsymbol{S}_{t}^{-1}, \\
\hat{\boldsymbol{x}}_{t} &=& \boldsymbol{K} \boldsymbol{z}_{t}+ \hat{\boldsymbol{x}}_{t \mid t-1}-\boldsymbol{K} h(\hat{\boldsymbol{x}}_{t \mid t-1}),\\  
\boldsymbol{P}_{t} &=& (\boldsymbol{I}-\boldsymbol{K}\boldsymbol{J}_{h}(\hat{\boldsymbol{x}}_{t \mid t-1}))\boldsymbol{P}_{t \mid t-1},
\end{eqnarray}
where $h(\cdot)$ is the nonlinear measurement function and $\boldsymbol{J}_{h}(\cdot)$ refers to the Jacobian of $h(\cdot)$. Here, we also consider a quadratic function as the nonlinear equation, i.e., $h(x) = \beta_0 + \beta_1 x + \beta_2 x^2$, where $\beta_0,\beta_1,\beta_2$ can be obtained by a neural network.
\end{itemize}

\textbf {Decoder}:
The output of Kalman filter $\boldsymbol{o}^{kal}_{t} \in \mathbb{R}^D$ at time $t$ can be achieved by using (20), 
\begin{equation}
\boldsymbol{o}^{kal}_{t}=Relu(\boldsymbol{W}_{k}^{\top} \boldsymbol{\hat{x}}_t + \boldsymbol{b}_{k}),
\end{equation}
where $\boldsymbol{W}_{k}$ and $\boldsymbol{b}_{k}$ are learnable parameters, respectively.

Finally, the exogenous information is further fused to enhance prediction accuracy. To be specific, let $\boldsymbol{W}_1,\boldsymbol{W}_2,\boldsymbol{W}_3,\boldsymbol{b}$ denote learnable parameters, then the prediction value $\boldsymbol{\hat{Y}}$ can be obtained with the following formulation,
\begin{equation}
\hat{\boldsymbol{Y}} = Relu(\boldsymbol{W}^{\top}_1 \boldsymbol{o}^{e_2}_t + \boldsymbol{W}^{\top}_2 \boldsymbol o_{t}^{ar} + \boldsymbol{W}^{\top}_3\boldsymbol{o}^{kal}_t + \boldsymbol{b}).
\end{equation}

\section{Experiments}
\subsection{Dataset Description}
We conduct experiments on three real-world datasets, including telco data from the city of Milan and the province of Trentino as well as LTE traffic data from a private operator. 
				
The telco data\cite{Barlacchi2015} record three types of call detail records (CDRs), i.e., the short message service (SMS), call service, and Internet service from 1 November 2013 to 1 January 2014 with a temporal interval of 10 minutes. For convenience, we concentrate on Internet traffic forecasting with time granularity in hours, and our proposed models can also be applied to the prediction of SMS and call services. Regarding the spatial dimension, Milan city is divided into a grid of (100$\times$100) squares, whereas the Trentino region is partitioned into a grid of (117$\times$98) cells. They are publicly available and can be obtained on Harvard Dataverse\cite{milan_data,trentino_data}. 
		
In terms of the auxiliary information in Milan and Trentino, we consider two sources of exogenous data, including social pulse and website news data. The social pulse data are composed of Twitter messages posted by users, which contain the anonymized username, geographical location, and the timestamp of the published tweet, etc. Moreover, the news data collect published articles on the websites, containing the topics and types of published articles as well as the geographical locations. Both datasets are also publicly available. Specifically, the social pulse and news data of Milan city can be obtained through\cite{Social_Pulse_milan,milanotoday}, and that of Trentino province can be accessed via\cite{Social_Pulse_trentino,TrentoToday}.
		
The LTE traffic data are collected from a private operator. It records the downlink traffic volume (GB) from 28 December 2020 to 12 January 2021 with a temporal interval of 1 hour. Besides, we use the historical downlink average throughput data and the timestamp information as auxiliary factors. This dataset is private and not available to the general public.

\subsection{Evaluation Metrics}
We choose root mean square error (RMSE), mean absolute error (MAE), and empirical correlation coefficient (CORR)\cite{b37} as evaluation metrics to assess the performance of different methods for cellular traffic prediction. They are defined as, 
\begin{eqnarray}
RMSE & = & \sqrt{\frac{1}{N}\sum_{i=1}^{N}(\boldsymbol{Y}_i-\hat{\boldsymbol{Y}_i})^{2}}, \\
MAE & = &\frac{1}{N}\sum_{i=1}^{N}|\boldsymbol{Y}_i-\hat{\boldsymbol{Y}_i}|, 
\end{eqnarray}
\begin{eqnarray}	
\boldsymbol{D}_{i,d} & = & \boldsymbol{Y}_{i,d}-mean(\boldsymbol{Y}_i),\\
\hat{\boldsymbol{D}}_{i,d} & = & \hat{\boldsymbol{Y}}_{i,d}-mean(\hat{\boldsymbol{Y}_i}),\\
CORR & = & \frac{1}{N}\sum_{i=1}^N\frac{\sum_{d=1}^D(\boldsymbol{D}_{i,d})(\hat{\boldsymbol{D}}_{i,d})}{\sqrt{\sum_{d=1}^D(\boldsymbol{D}_{i,d})^2(\hat{\boldsymbol{D}}_{i,d})^2}}.
\end{eqnarray}

\subsection{Baselines}
To demonstrate the effectiveness of the proposed models, we compare A-LKF and A-EKF with the following baselines.
\begin{itemize}
\item \textbf {LSTM}: It is a deep learning method using LSTM cells that can learn both short-term and long-term temporal dependencies within time series.
\item \textbf {GRU}: This is a sequential model using GRU cells, which is of a simple structure and has been widely used in time series prediction.
\item \textbf{GCN}: This architecture is utilized to capture the spatial dependencies of traffic data.
\item \textbf {DeseNet}\cite{b25}: It utilizes paralleled CNN to adaptively capture spatiotemporal correlations of cellular traffic among neighboring cells. 
\item \textbf {DeepAuto}\cite{b32}: It uses three paralleled LSTM to capture the trend, period, and closeness traffic patterns. Besides, auxiliary factors have been utilized to improve the prediction performance.
\item \textbf {STCNet}\cite{b31}: It uses convolutional long short term memory (ConvLSTM) and CNN to model the spatiotemporal dependencies and exogenous factors, respectively.
\item \textbf {ST-Tran}\cite{add-ref2}: It utilizes transformer to characterize the spatial-temporal relationships of cellular traffic.
\end{itemize}

\subsection{Experimental Details}
\textbf{Preprocessing:} 
The Min-Max normalization\cite{b31} is utilized to scale the numerical data such as cellular traffic, tweet coordinates, and the number of users, tweets, and articles to the interval of [0, 1]. Furthermore, we transform metadata such as holidays and the day of the week via one-hot encoding.

\textbf{Hyperparameters:} The grid search is conducted for all tunable hyperparameters. For other models, all baseline models use the parameters provided by the papers. Specifically, we select RMSE as the loss objective function and utilize Adam algorithm\cite{b40} to optimize parameters.

In our study, we select the neighboring 25 cells for experiment with the center cell in the middle. In terms of the center cells, we choose the $square\_id$ of $5060$ $(W=50,H=60)$, $4259$ $(W=42,H=59)$ in Milan and the $square\_id$ of $5085$ $(W=50,H=85)$, $5680$ $(W=56,H=80)$ in Trentino. We set horizon $h = \{1, 24\}$ to forecast traffic for the next 1 hour and 1 day, respectively. We reckon cellular traffic from the last seven days as the test data and all traffic before as training data and validate data. The optimal results are highlighted in boldface and underlined italics for each metric. We conduct all experiments on a Linux server with four 12GB GPUs with NVIDIA TITAN X (Pascal).

\section{Results}
In this section, we conduct extensive experiments on three real-world datasets to evaluate the performance of a total of nine methods for cellular traffic prediction.

\begin{table*}[t]
\centering
\caption{\MakeUppercase{experiment results over three real-world datasets}} 
\label{table}
\useunder{\uline}{\ul}{}
\setlength{\tabcolsep}{3pt}
\resizebox{\textwidth}{!}{
\begin{tabular}{c|c|cccc|cccc|cc|c}
\hline
\multirow{3}{*}{Methods}  & \multirow{3}{*}{Metrics} & \multicolumn{4}{c|}{Milan}                                                                                                             & \multicolumn{4}{c|}{Trentino}                                                                                                       & \multicolumn{2}{c|}{\multirow{2}{*}{LTE traffic}}                  & \multirow{3}{*}{Average rank} \\ \cline{3-10}
                          &                          & \multicolumn{2}{c}{5060}                                           & \multicolumn{2}{c|}{4259}                                         & \multicolumn{2}{c}{5680}                                        & \multicolumn{2}{c|}{5085}                                         & \multicolumn{2}{c|}{}                                         &                               \\ \cline{3-12}
                          &                          & 1-hour                          & 1-day                            & 1-hour                          & 1-day                           & 1-hour                         & 1-day                          & 1-hour                          & 1-day                           & 1-hour                        & 1-day                         &                               \\ \hline
\multirow{3}{*}{LSTM}     & RMSE                     & 714.989                         & 2437.325                         & 285.552                         & 603.014                         & 26.235                         & 43.490                         & 178.569                         & 614.930                         & 1.378                         & 1.177                         & \multirow{3}{*}{6.30}         \\
                          & MAE                      & 491.200                         & 1729.045                         & 190.309                         & 430.330                         & 12.897                         & 16.837                         & 97.378                          & 188.738                         & 1.005                         & 0.894                         &                               \\
                          & CORR                     & 0.976                           & 0.757                            & 0.922                           & 0.780                           & 0.972                          & 0.932                          & 0.993                           & 0.918                           & 0.824                         & 0.872                         &                               \\ \hline
\multirow{3}{*}{GRU}      & RMSE                     & 657.876                         & 2416.961                         & 264.142                         & 639.602                         & 33.312                         & 42.964                         & 190.145                         & 477.463                         & 1.329                         & 1.183                         & \multirow{3}{*}{6.10}         \\
                          & MAE                      & 462.714                         & 1801.244                         & 176.150                         & 464.667                         & 15.867                         & 17.688                         & 100.925                         & 167.051                         & 1.005                         & 0.894                         &                               \\
                          & CORR                     & 0.979                           & 0.750                            & 0.930                           & 0.750                           & 0.957                          & 0.933                          & 0.992                           & 0.966                           & 0.849                         & 0.873                         &                               \\ \hline
\multirow{3}{*}{GCN}      & RMSE                     & 616.077                         & 2139.964                         & 235.415                         & 391.325                         & 34.306                         & 52.028                         & 153.923                         & 365.022                         & 1.067                         & 1.236                         & \multirow{3}{*}{7.30}         \\
                          & MAE                      & 389.240                         & 1359.980                         & 158.398                         & 263.190                         & 15.429                         & 21.754                         & 68.259                          & 133.708                         & 0.790                         & 0.934                         &                               \\
                          & CORR                     & 0.965                           & 0.552                            & 0.872                           & 0.692                           & 0.948                          & 0.903                          & 0.995                           & 0.973                           & 0.898                         & 0.862                         &                               \\ \hline
\multirow{3}{*}{DenseNet} & RMSE                     & 654.527                         & 2102.087                         & 167.468                         & 403.413                         & 22.929                         & 42.420                         & 148.901                         & 276.882                         & 1.016                         & 1.074                         & \multirow{3}{*}{5.00}         \\
                          & MAE                      & 483.256                         & 1106.953                         & 126.054                         & 284.166                         & 11.585                         & 19.848                         & 69.723                          & 91.128                          & 0.757                         & 0.829                         &                               \\
                          & CORR                     & 0.965                           & 0.613                            & 0.926                           & 0.667                           & 0.977                          & 0.928                          & 0.993                           & 0.979                           & 0.906                         & 0.898                         &                               \\ \hline
\multirow{3}{*}{DeepAuto} & RMSE                     & 673.683                         & 1747.707                         & 214.453                         & 324.884                         & 24.140                         & 36.175                         & 154.243                         & 322.751                         & 0.944                         & 1.101                         & \multirow{3}{*}{5.07}         \\
                          & MAE                      & 442.837                         & 1062.018                         & 138.126                         & 219.375                         & 12.170                         & 15.754                         & 61.717                          & 104.442                         & 0.733                         & 0.836                         &                               \\
                          & CORR                     & 0.956                           & 0.693                            & 0.886                           & 0.741                           & 0.973                          & 0.944                          & 0.993                           & 0.967                           & 0.919                         & 0.890                         &                               \\ \hline
\multirow{3}{*}{STCNet}   & RMSE                     & 537.828                         & 1803.966                         & 161.626                         & 399.160                         & 25.140                         & 42.259                         & 132.331                         & 280.397                         & 1.009                         & 1.108                         & \multirow{3}{*}{4.30}         \\
                          & MAE                      & 367.181                         & 1107.565                         & 123.967                         & 267.820                         & 13.785                         & 21.038                         & {\ul \textit{\textbf{55.359}}}  & 96.085                          & 0.768                         & 0.872                         &                               \\
                          & CORR                     & 0.974                           & 0.813                            & 0.924                           & 0.729                           & 0.973                          & 0.929                          & 0.996                           & 0.982                           & 0.907                         & 0.890                         &                               \\ \hline
\multirow{3}{*}{ST-Trans} & RMSE                     & 560.002                         & 1749.913                         & {\ul \textit{\textbf{135.399}}} & 325.669                         & {\ul \textit{\textbf{21.080}}} & 38.639                         & {\ul \textit{\textbf{122.243}}} & 270.786                         & {\ul \textit{\textbf{0.935}}} & 1.069                         & \multirow{3}{*}{2.67}         \\
                          & MAE                      & 349.700                         & {\ul \textit{\textbf{978.282}}}  & {\ul \textit{\textbf{94.601}}}  & 239.818                         & {\ul \textit{\textbf{8.428}}}  & 16.760                         & 61.511                          & 119.257                         & {\ul \textit{\textbf{0.702}}} & 0.827                         &                               \\
                          & CORR                     & 0.976                           & 0.762                            & 0.951                           & 0.800                           & 0.979                          & 0.933                          & 0.996                           & 0.978                           & {\ul \textit{\textbf{0.922}}} & 0.897                         &                               \\ \hline
\multirow{3}{*}{A-LKF}    & RMSE                     & {\ul \textit{\textbf{536.854}}} & {\ul \textit{\textbf{1671.606}}} & 183.525                         & 336.702                         & 21.179                         & {\ul \textit{\textbf{35.060}}} & 124.239                         & 257.117                         & 0.937                         & 1.008                         & \multirow{3}{*}{2.03}         \\
                          & MAE                      & {\ul \textit{\textbf{341.316}}} & 1013.801                         & 124.899                         & {\ul \textit{\textbf{227.735}}} & 10.535                         & 15.556                         & 59.168                          & 107.924                         & 0.704                         & 0.765                         &                               \\
                          & CORR                     & {\ul \textit{\textbf{0.985}}}   & {\ul \textit{\textbf{0.856}}}    & {\ul \textit{\textbf{0.962}}}   & {\ul \textit{\textbf{0.892}}}   & {\ul \textit{\textbf{0.982}}}  & {\ul \textit{\textbf{0.955}}}  & 0.996                           & 0.984                           & 0.918                         & 0.906                         &                               \\ \hline
\multirow{3}{*}{A-EKF}    & RMSE                     & 558.668                         & 1676.569                         & 185.542                         & {\ul \textit{\textbf{323.703}}} & 21.093                         & 36.691                         & 124.239                         & {\ul \textit{\textbf{255.563}}} & 0.937                         & {\ul \textit{\textbf{0.998}}} & \multirow{3}{*}{2.00}         \\
                          & MAE                      & 348.675                         & 1037.913                         & 125.340                         & 235.732                         & 10.419                         & {\ul \textit{\textbf{14.414}}} & 59.168                          & {\ul \textit{\textbf{83.772}}}  & 0.712                         & {\ul \textit{\textbf{0.752}}} &                               \\
                          & CORR                     & 0.983                           & 0.856                            & 0.959                           & 0.876                           & 0.980                          & 0.953                          & {\ul \textit{\textbf{0.996}}}   & {\ul \textit{\textbf{0.985}}}   & 0.918                         & {\ul \textit{\textbf{0.909}}} &                               \\ \hline
\end{tabular}}
\end{table*}

\subsection{Prediction Performance and Analysis}
In this section, we rank the results for each group and utilize the average rank to compare the prediction performance. Table I summarizes the results on three real-world datasets with RMSE, MAE, and CORR metrics. It is observed that LSTM, GRU neural networks, and GCN perform poorly among the nine approaches considered in our study. It is because LSTM and GRU neural networks cannot capture the spatial dependencies, while GCN cannot effectively extract the long-term temporal dependencies of cellular traffic data. 
DeseNet can extract cellular traffic's closeness, period, and trend patterns with three paralleled CNN, thus performing better than LSTM neural networks, GRU neural networks, and GCN. For DeepAuto, the three LSTM neural networks are adopted to extract the instantaneous, periodic, and seasonal patterns, respectively. Besides, the external factors are also merged with a fusion layer. Therefore, DeepAuto performs better than LSTM and GRU neural networks. For STCNet, ConvLSTM and CNN are separately applied to seize the spatiotemporal dependencies and exogenous features, verifying the validity of auxiliary features for cellular traffic prediction. 

ST-Tran\cite{add-ref2} is a spatiotemporal prediction model based on transformer. It can be observed that ST-Tran achieves the best performance among the above baselines. Compared with our proposed methods (i.e., A-LKF and A-EKF), it performs better in the 1-hour ahead prediction but worse in the 1-day ahead prediction. This is because that transformer has the error accumulation problem when performing long-term time series prediction\cite{bohu2022}. As the timestep increases, small errors gradually lead to large errors. In contrast, our proposed models can use Kalman gain to adaptively balance the observation and priori state to obtain the optimal posteriori state at each timestep, allowing the accumulated errors in the observations to be alleviated. Therefore, our proposed models can provide more accurate predictions in long-term traffic forecasting. 

The performance of our proposed models, including A-LKF and A-EKF, generally outperform the above benchmarks. There are several possible explanations for this. \textit{Firstly}, we extract the auxiliary features from multi-source heterogeneous data for cellular traffic prediction, contributing to
improving the prediction accuracy. \textit{Secondly}, the encoder module and Kalman filter can fully extract the spatiotemporal correlations of cellular traffic. In particular, the Kalman gain is used to adaptively balance the observation and the priori state at each timestep, thereby reducing the error accumulation and enhancing the prediction performance. \textit{Thirdly}, AR can capture the scale changes of traffic and further improve the prediction accuracy of our proposed models.

Furthermore, in terms of the two proposed variants, A-EKF slightly performs better than A-LKF. It is known that traffic data have nonlinear characteristics. A-LKF adopts linear state and measurement equations to capture the long-term temporal dependencies. In comparison, A-EKF uses Taylor series expansion to approximate nonlinear functions, which can more accurately model the nonlinear temporal characteristics of the traffic sequence, Thus obtaining better prediction performance than A-LKF.

\subsection{Ablation Study}
To better understand the importance of each module (i.e., the attention mechanism, the exogenous feature extraction, and AR), we conduct the ablation experiment on LTE traffic data. Specifically, we only remove one module in A-LKF framework and name A-LKF without different structures.
\begin{itemize}
\item \textbf{A-LKF/oAr:} The A-LKF architecture without AR.
\item \textbf{A-LKF/oExo:} The A-LKF architecture without the exogenous feature extraction.
\item \textbf{A-LKF/oAtt:} The A-LKF architecture without the attention mechanism.
\end{itemize}

\begin{figure}[htb]
\begin{center}
    \centerline{\includegraphics[width=0.8\columnwidth]{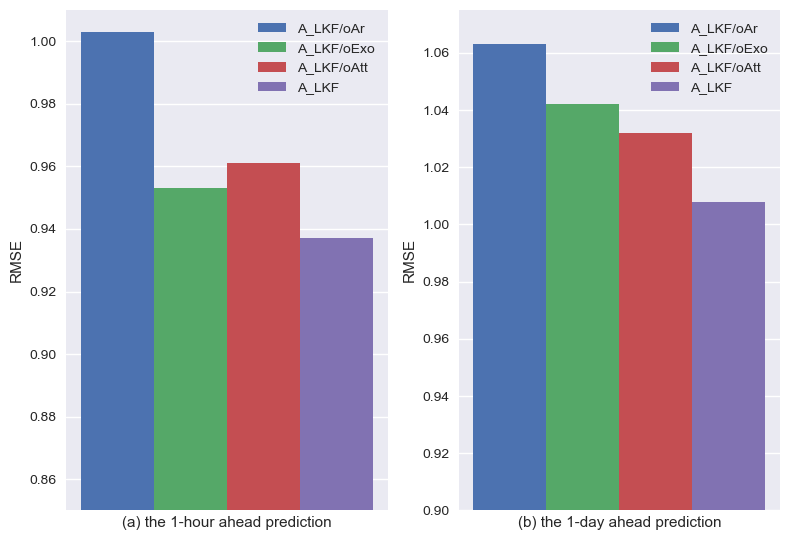}}
Fig. 2. The prediction performance in terms of RMSE.
\label{Fig}
\end{center}
\end{figure}

\begin{figure}[htb]
\begin{center}
    \centerline{\includegraphics[width=0.8\columnwidth]{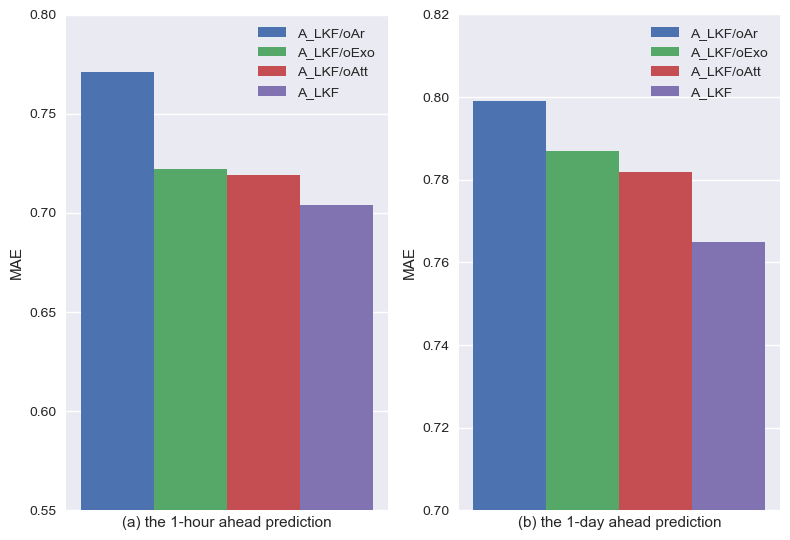}}
Fig. 3. The prediction performance in terms of MAE.
\label{Fig}
\end{center}
\end{figure}

\begin{figure}[htb]
\begin{center}
    \centerline{\includegraphics[width=0.8\columnwidth]{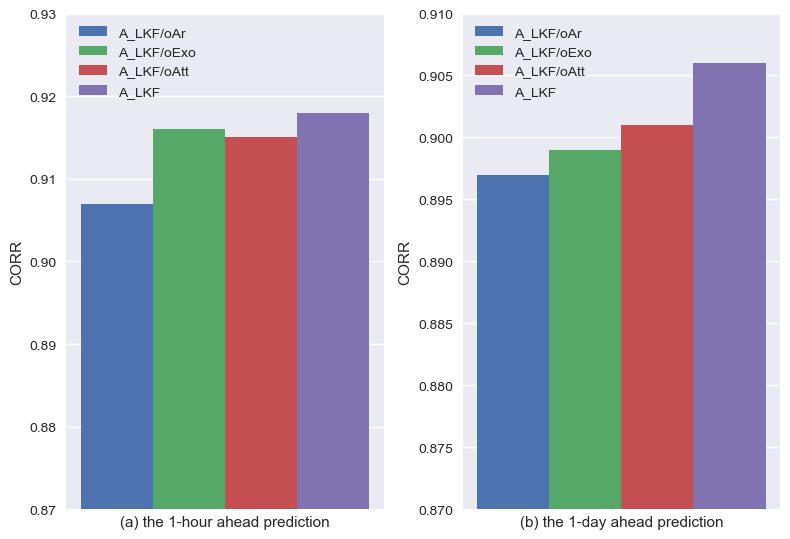}}
	Fig. 4. The prediction performance in terms of CORR.
	\label{Fig}
\end{center}
\end{figure}

The results of the ablation experiment are displayed in Fig. 2, Fig. 3, and Fig. 4 in terms of RMSE, MAE, and CORR, respectively. It can be observed that removing the attention mechanism or exogenous feature extraction (i.e., A-LKF/oAtt or A-LKF/oExo) indeed deteriorates the prediction performance to some degree, but removing AR (i.e., A-LKF/oAr) has the most significant impact on performance degradation. Collectively, the results show that these three modules together lead to the better predictive performance of our proposed models for cellular traffic prediction.

\subsection{Visualization Analysis}
Some related studies reveal that Kalman filter is interpretable as it yields valuable interpretable information\cite{Revach2022_KalmanNet,Zhao2021_Interpretable}. For instance, the Kalman gain is an adaptive weight that can balance the observation and priori state to obtain the optimal posteriori state at each timestep. 

For better illustration, we visualize the Kalman gain in A-LKF over 24 hours, which can be seen in Fig. 5. The mean of the Kalman gain matrix is calculated and averaged over all sequences, which can be used as an aggregated indicator of the Kalman gain matrix. It can be observed that when timestep is less than six, the indicator decreases with the timestep increases. It shows that the model relies more on the observation. As timestep continues to increase, the proportion of the priori state gradually increases. When timestep reaches 15, the Kalman gain tends to be stable, indicating that the observation and priori state have reached equilibrium. Therefore, the visualization of the Kalman gain can provide an excellent opportunity to explain how the system state changes over time when extracting the long-term temporal dependencies.

\begin{figure}[htb]
\begin{center}
	\centerline{\includegraphics[width=0.7\columnwidth]{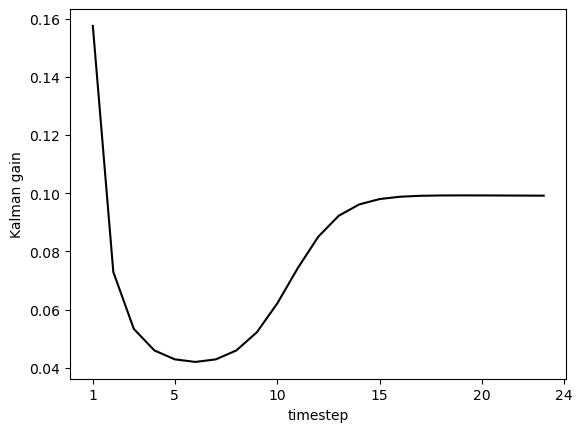}}
	 Fig. 5. The visualization of the Kalman gain over time.
	 \label{Fig}
\end{center}
\end{figure}

\section{Conclusion}
This paper proposes a DSSM with two variants, including A-LKF and A-EKF, to capture spatiotemporal dependencies of cellular traffic among neighbouring cells. Moreover, our proposed models consist of five modules: an encoder, Kalman filter, a decoder, exogenous feature extraction, and AR. In particular, an encoder is implemented to extract the spatial dependencies, and Kalman filter is employed to capture the long-term temporal correlations. Moreover, exogenous feature extraction and AR are also adopted to improve predictive performance accuracy further. Extensive experiments have been conducted on three real-world datasets. The results show the effectiveness of our proposed models and outperform
other state-of-the-art approaches for cellular traffic forecasting. Besides, the ablation experiment reveals that the attention mechanism, the exogenous feature extraction, and AR are jointly utilized to enhance the prediction performance.

One possible extension of this paper is to utilize a knowledge graph to extract exogenous factors from semantic information for cellular traffic prediction. Besides, the estimation of the dynamic systems can also be extended to unscented Kalman filter and particle filter, which would be an exciting direction for future research.

\bibliographystyle{IEEEtran}
\bibliography{References}

\begin{IEEEbiography}
[{\includegraphics[width=1in,height=1.25in,clip,keepaspectratio]{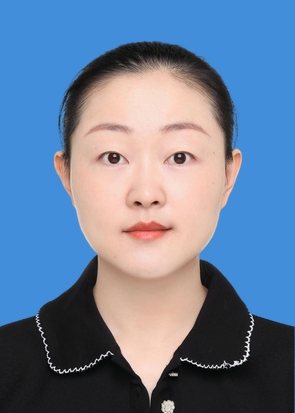}}]
{Hui Ma}received her B.Eng. degree in 2014 and M.S. degree in 2017 from Jiangnan University, Wuxi, China. She is a Ph.D. candidate in the Department of Computer Science, Tongji University, Shanghai, China. Her current research interests include deep learning and time series forecasting.
\end{IEEEbiography}

\begin{IEEEbiography}[{\includegraphics[width=1in,height=1.25in,clip,keepaspectratio]{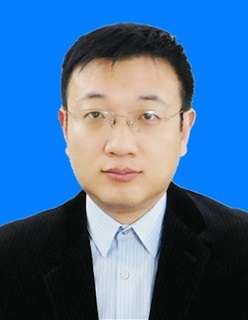}}]
{Kai Yang} received the B.Eng. degree from Southeast University, Nanjing, China, the M.S. degree from the National University of Singapore, Singapore, and the Ph.D. degree from Columbia
University, New York, NY, USA.

He is a Distinguished Professor with Tongji University, Shanghai, China. He was a Technical
Staff Member with Bell Laboratories, Murray Hill, NJ, USA. He has also been an Adjunct Faculty
Member with Columbia University since 2011. He holds over 20 patents and has been published extensively in leading IEEE journals and conferences. His current research interests include big data analytics, machine learning, wireless communications, and signal processing.
\end{IEEEbiography}

\begin{IEEEbiography}
[{\includegraphics[width=1in,height=1.25in,clip,keepaspectratio]{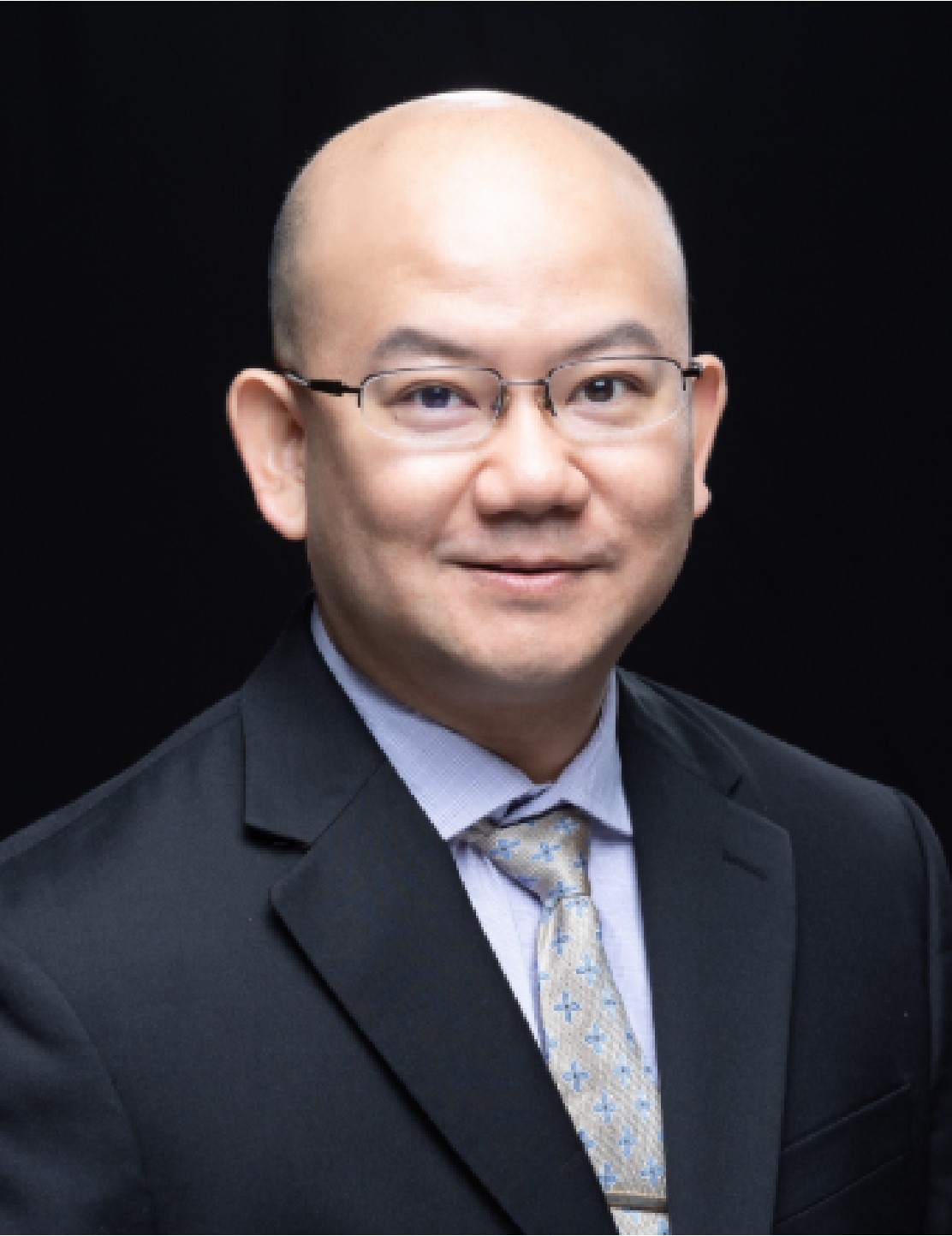}}]
{Man-On Pun}(Senior Member, IEEE) received the Ph.D. degree in electrical engineering from the University of Southern California (USC), Los Angeles, USA, in 2006. He was a Postdoctoral Research Associate with Princeton University from 2006 to 2008. He is currently an Associate Professor with the School of Science and Engineering, The Chinese University of Hong Kong at Shenzhen (CUHKSZ), Shenzhen. Prior to joining CUHKSZ in 2015, he held research positions at Huawei (USA), Mitsubishi Electric Research Labs (MERL), Boston and Sony, Tokyo, Japan. Prof. Pun’s research interests include AI Internet of Things (AIoT) and applications of machine learning in communications and satellite remote sensing.

\end{IEEEbiography}

\end{document}